\documentclass[conference]{IEEEtran}
\IEEEoverridecommandlockouts

\usepackage{cite}
\usepackage{amsmath,amssymb,amsfonts}
\usepackage{mathrsfs}
\usepackage{graphicx}
\usepackage{textcomp}
\usepackage{xcolor}
\usepackage{balance}

\usepackage[compact]{titlesec} 

\usepackage[ruled,vlined,linesnumbered]{algorithm2e}
\usepackage{amsmath}
\usepackage{graphicx, wrapfig, subcaption, setspace, booktabs}

\newtheorem{proposition}{\textbf{Proposition}}

\newtheorem{definition}{\textbf{Definition}}

\makeatletter

\def\BibTeX{{\rm B\kern-.05em{\sc i\kern-.025em b}\kern-.08em
    T\kern-.1667em\lower.7ex\hbox{E}\kern-.125emX}}
\begin{document}

\title{Scalable Deep Subspace Clustering Network}

\author{\IEEEauthorblockN{Nairouz Mrabah} 
\IEEEauthorblockA{\textit{Department of Computer Science} \\
\textit{University of Quebec at Montreal}\\
Montreal, Quebec, Canada \\
mrabah.nairouz@courrier.uqam.ca}
\and
\IEEEauthorblockN{Mohamed Bouguessa} 
\IEEEauthorblockA{\textit{Department of Computer Science} \\
\textit{University of Quebec at Montreal}\\
Montreal, Quebec, Canada \\
bouguessa.mohamed@uqam.ca}
\and
\IEEEauthorblockN{Sihem Sami} 
\IEEEauthorblockA{\textit{Department of Computer Science} \\
\textit{University of Quebec at Montreal}\\
Montreal, Quebec, Canada \\
sami.sihem@courrier.uqam.ca}

}

\maketitle

\newcommand{\SDSNet}{SDSNet}
\begin{abstract}
Subspace clustering methods face inherent scalability limits due to the $O(n^3)$ cost (with $n$ denoting the number of data samples) of constructing full $n\times n$ affinities and performing spectral decomposition. While deep learning-based approaches improve feature extraction, they maintain this computational bottleneck through exhaustive pairwise similarity computations. We propose SDSNet (Scalable Deep Subspace Network), a deep subspace clustering framework that achieves $\mathcal{O}(n)$ complexity through (1) landmark-based approximation, avoiding full affinity matrices, (2) joint optimization of auto-encoder reconstruction with self-expression objectives, and (3) direct spectral clustering on factorized representations. The framework combines convolutional auto-encoders with subspace-preserving constraints. Experimental results demonstrate that SDSNet achieves comparable clustering quality to state-of-the-art methods with significantly improved computational efficiency.
\end{abstract}


\section{Introduction}

The rapid expansion of high-dimensional data has intensified the demand for effective and scalable clustering methods. 
Subspace clustering defines a cluster as a set of points lying in the same low-dimensional linear subspace. This formulation is well-suited for high-dimensional datasets where multiple subspaces coexist. It has found widespread application in areas such as image segmentation \cite{paper2}, motion analysis \cite{paper3, paper5}, and face grouping \cite{paper5, paper4}, where the subspace assumption aligns naturally with the data's intrinsic structure.

A significant advancement in subspace clustering comes from methods leveraging the self-expressiveness property \cite{paper6, paper9, paper8}, which assumes that each data point can be expressed as a linear combination of other points within the same subspace. Sparse Subspace Clustering (SSC) \cite{paper6} and related approaches construct an affinity matrix by solving a sparse representation problem, followed by spectral clustering. While these methods have demonstrated competitive clustering performance, their computational complexity scales as \( \mathcal{O}(n^3) \) due to the need for affinity matrix construction and spectral decomposition.

Addressing the computational complexity of subspace clustering is a central challenge. Several strategies have been proposed, including sampling techniques, which assume that data contains redundant information, allowing a smaller subset of data points to approximate the original structure. Theoretically, for each linear subspace, a small set of independent samples, equal in size to the subspace dimensions, is sufficient to recover the clustering subspaces.

Sampling-based approaches can follow a targeted strategy, selecting representative points based on optimization criteria, or a random strategy, selecting points without constraints. Targeted methods \cite{paper16, paper38, paper30} improve subspace representation but may incur high computational costs due to iterative optimization. Moreover, they often assume independent subspaces, a restrictive condition rarely satisfied in real-world datasets where subspaces may overlap \cite{paper41}. Random selection techniques \cite{paper15, paper25} avoid costly optimization but might fail to guarantee sufficient subspace coverage and thus degrade clustering performance. Other approaches \cite{paper42, paper19, paper18, paper22} aim to improve affinity graph construction while preserving subspace structures. These methods typically impose theoretical constraints that may not generalize well to real-world datasets. 

From another perspective, subspace clustering models are only suitable for data points that are linearly spread within subspaces. Unfortunately, such an assumption can hardly be met in general cases due to noise and outliers. To make subspace clustering models reliable under general circumstances, the raw data must first be transformed into a representation that alleviates noise while still revealing the underlying subspace structure. Auto-encoders, in particular, have shown success in extracting robust latent representations that can capture the underlying subspaces \cite{paper28, paper43}. These methods introduce a self-expression layer between the encoder and decoder to construct an affinity matrix in the latent space.

However, the majority of deep subspace clustering methods still rely on full pairwise affinity matrix construction and spectral clustering, making them computationally expensive. A notable scalable approach, Scalable Deep \( k \)-Subspace Clustering \cite{paper30}, eliminates the affinity matrix learning step and adapts the idea of k-subspace clustering \cite{paper46} in the latent space of a deep auto-encoder network. Consequently, this method reduces the computational complexity to \( \mathcal{O}(n^2) \). While this improves efficiency, it still requires costly SVD computations.

Few other methods~\cite{paper39, paper44} have tried to relax the heavy $\mathcal{O}(n^{3})$ cost of classical deep subspace clustering networks. The model of~\cite{paper39}—Deep Low-Rank Subspace Clustering (DLRSC)—shrinks the complexity for computing the $n\times n$ self-expression matrix by a rank-$m$ factorisation ($m < n$) and adds a nuclear-norm penalty that is evaluated on the much smaller $m\times m$ core. Nevertheless, DLRSC still computes the full affinity matrix after training the auto-encoder and runs standard spectral clustering on the resulting $n\times n$ Laplacian. This leaves its overall complexity at a cubic cost.

The more recent—Efficient Deep Embedded Subspace Clustering (EDESC)—of ~\cite{paper44} goes further by removing the self-expression layer. Because no $n\times n$ affinity matrix or eigen-decomposition is required, EDESC achieves linear computational complexity. The trade-off is that it assumes each point belongs to a single subspace, so performance may degrade when subspaces overlap or $k$ is misspecified. A key advantage of self–expression–based methods is that, under mild conditions, the learned self-expression matrix is subspace preserving: non–zero connections appear only between points that belong to the same subspace, which leads to provable recovery guarantees. EDESC abandons these guarantees.

To eliminate the cubic bottleneck of deep subspace clustering, we introduce \emph{Scalable Deep Subspace Network (SDSNet)}, the first self-expression model whose entire pipeline—affinity construction and spectral assignment—runs in time proportional to the number of samples \(n\). SDSNet replaces the dense self-expression matrix with a low-rank factorisation and a small set of landmark points, producing a compact affinity matrix that is guaranteed to be symmetric and positive-semidefinite. The associated graph Laplacian lives in the same low-dimensional subspace as the factors, so its leading eigenvectors are recovered from a tiny eigenproblem rather than a full \(n \times n\) decomposition. Coupled with a convolutional auto-encoder for robust feature extraction, SDSNet achieves competitive clustering quality while retaining linear complexity. The primary contributions of this work are as follows:

\noindent \textbf{(1) A computationally efficient subspace clustering algorithm:} We propose an approach that reduces the complexity of self-expression matrix construction and spectral clustering from \( \mathcal{O}(n^3) \) to a linear complexity of \( \mathcal{O}(n) \). This significantly reduces the computational burden of subspace clustering. \\ \textbf{(2) Convergence towards optimal subspaces:} Our method selects a subset of representative landmark points, considerably smaller than the original dataset, to learn similarity relationships between data points. This enables computational efficiency while maintaining clustering accuracy.\\ \textbf{(3) A novel deep learning-based subspace clustering approach:} Our method leverages auto-encoders with a self-expression layer to handle complex data that may not inherently reside in linear subspaces. We define a new objective function to further optimize the subspace clustering process while reducing computational complexity.\\ \textbf{(4) Comprehensive experimental validation:} We evaluate our method on both synthetic and real-world datasets. The results demonstrate a significant improvement in clustering accuracy compared to existing scalable solutions. Furthermore, experiments confirm that our approach effectively identifies subspaces, producing results that remain competitive with deep subspace clustering methods that rely on full pairwise similarity computations.

\section{Related Work}
\label{sec:related}

In alignment with the goals of this work, we examine two prominent methodologies of subspace clustering: scalable subspace clustering and deep subspace clustering. Scalable methods focus on improving computational efficiency, whereas deep learning-based approaches enhance representation learning through more expressive models. In this section, we review key contributions from both directions and emphasize the importance of developing a method that leverages the strengths of both approaches while mitigating their weaknesses.

\textbf{Scalable Subspace Clustering.} Large-Scale Multi-View Subspace Clustering in Linear Time (LMVSC) \cite{paper15} constructs an anchor graph instead of a full self-expression matrix, significantly reducing computational complexity. A subset of $m$ ($m \ll n$) anchor points is selected via k-means, and each data point is expressed as a linear combination of anchors within the same subspace. To further optimize efficiency, LMVSC replaces spectral clustering with an equivalent linear-time matrix factorization strategy \cite{paper35}. However, it lacks theoretical guarantees for recovering all data subspaces and is sensitive to noise and outliers, limiting clustering accuracy. The exact subspace clustering in linear time method \cite{paper16} follows a sampling-clustering-classification framework. It first selects a minimal number of anchor points—equal to the rank of the data matrix—using a principled technique that ensures independence, theoretically guaranteeing subspace recovery in linear time. Standard subspace clustering methods (e.g., SSC, LRR) are then applied, followed by a regression-based out-of-sample extension. While theoretically sound, this approach struggles with missing values, outliers, and noise variations, impacting clustering performance.

Sparse Subspace Clustering via Orthogonal Matching Pursuit (SSC-OMP) \cite{paper17} uses Orthogonal Matching Pursuit (OMP) to obtain sparse self-expression coefficients. While it provides subspace-preserving guarantees, its quadratic complexity makes it slower than nuclear norm-based methods (e.g., LRR, LSR). Additionally, its sparsity constraint weakens connectivity within subspaces, increasing the risk of over-segmentation \cite{paper18} and reducing clustering accuracy. Oracle-Based Active Set Algorithm for Elastic Net Subspace Clustering \cite{paper19} explores the elastic net regularizer (a combination of $\ell_1$ and $\ell_2$ penalties) to balance subspace preservation and connectivity. Computational efficiency is improved by solving subproblems on an incrementally growing subset of data. However, the method does not achieve linear complexity and remains costly even for medium-sized datasets. Finally, the Stochastic Sparse Subspace Clustering (SSSC) approach \cite{paper18} introduces Dropout to mitigate over-segmentation in sparse subspace clustering, effectively adding $\ell_2$ regularization to the least squares regression model. However, random dropout does not guarantee optimal within-cluster connectivity, limiting clustering accuracy. Moreover, the reliance on spectral clustering introduces a cubic-time complexity bottleneck.

\textbf{Deep Subspace Clustering.} 
Deep Subspace Clustering Network (DSCNet) \cite{paper28} applies an auto-encoder with a self-expression layer to enforce subspace-preserving representations. Although effective, its cubic complexity makes it impractical for large datasets. 
Deep Adversarial Subspace Clustering (DASC) \cite{paper29} extends DSCNet with adversarial training. This approach improves clustering accuracy but inherits the cubic complexity of DSCNet, which limits scalability. Pseudo-Supervised Deep Subspace Clustering (PSDSC) \cite{paper34} refines DSCNet by incorporating pseudo-supervision and a sampling–clustering–classification strategy. However, it lacks theoretical guarantees, relies on a small training subset, and does not define sample selection systematically, affecting clustering performance. Self-Supervised Convolutional Subspace Clustering Network (S$^2$ConvSCN) \cite{paper32} jointly optimizes spectral clustering and self-expression objectives to improve feature learning. 
Nevertheless, this significantly increases computational cost, making it unsuitable for large datasets. Deep Subspace Clustering with Data Augmentation \cite{paper33} improves clustering robustness by enforcing latent space consistency across augmentations. Although it enhances clustering accuracy, the method remains computationally expensive. Scalable Deep $k$-Subspace Clustering (kSCN) \cite{paper30} replaces the self-expression loss with a deep adaptation of $k$-subspace clustering \cite{paper31}. By directly constraining latent representations, it avoids affinity matrix construction. Although more scalable than prior deep models, its quadratic complexity and lower clustering accuracy remain limitations.

To conclude, scalable subspace clustering methods often struggle to achieve linear-time complexity while maintaining clustering accuracy, whereas deep subspace clustering methods suffer from high computational costs due to full pairwise affinity matrix construction. These challenges motivate the need for a subspace clustering approach that effectively balances computational efficiency and clustering accuracy. In the next section, we present our method, which leverages deep learning and structured approximations to reduce complexity while preserving clustering performance.

\section{Proposed Approach}

To overcome the cubic time bottleneck, 
we introduce SDSNet, a scalable deep subspace-clustering framework that achieves linear complexity with the number of samples while retaining competitive accuracy. Let $\mathbf{X}= \bigl[x_1,x_2,\ldots ,x_n\bigr]\in\mathbb{R}^{D\times n}$ denote the data matrix of \(n\) points in a space of dimension \(D\). 

We assume that the columns of \(\mathbf{X}\) lie in a union of \(k\) low-dimensional linear subspaces \(\{S_i\}_{i=1}^{k}\), which satisfy \(\sum_{i=1}^k D_i' \leq D\) and \(D_i'\) is the dimension of subspace $S_i$. The objective is to develop a scalable subspace clustering method that partitions the data into \(k\) clusters, where each cluster belongs to one or more subspaces. To achieve this, our approach reformulates the self-expressiveness objective of subspace clustering while eliminating the cubic terms found in prior work.  

Our pipeline first compresses the data through a convolutional auto-encoder that removes non-linear noise, then learns pair-wise affinities via a factorised self-expression matrix. The factorisation introduces \(m\) anchor embeddings (\(m \ll n\)) to reduce storage and computation from \(\mathcal{O}(n^{2})\) to \(\mathcal{O}(nm)\). This landmark-based step is optimised jointly with the reconstruction loss. The resulting factors feed an anchor-graph spectral routine whose eigen-problem is solved in the same low-dimensional space, yielding an overall clustering cost of \(\mathcal{O}(nm^{2})\). The subsections that follow detail feature extraction, scalable self-expressive modelling, optimisation, anchor-based spectral clustering, and complexity analysis.

\subsection{Feature Extraction}

We begin by describing the feature extraction process, which leverages a convolutional auto-encoder (CAE) to eliminate nonlinear noise and generate embeddings that better align with the subspace assumption. The encoding function $\mathscr{E}$ is a composition of $L$ convolutional layers, $\mathscr{E} = f^{(L)}\circ f^{(L-1)}\circ\cdots\circ f^{(1)}$ and each layer is defined by the function $f^{(L)}$ as follows:
\begin{equation}
    f^{(\ell)}(\mathbf{Z}^{(\ell)}) \,=\, \text{ReLU}\bigl(\mathbf{W}^{(\ell)} * \mathbf{Z}^{(\ell)} + \mathbf{b}^{(\ell)}\bigr), 
    \; \ell \in \{1,\dots,L\},
\end{equation}
\noindent where \(\text{ReLU}(x) = \max(0, \,x)\) is the rectified linear unit activation function; $\mathbf{W}^{(\ell)}$ and $\mathbf{b}^{(\ell)}$ are the convolution weight matrix and the bias vector, respectively, for the $\ell^{\text{th}}$ layer; “$*$’’ denotes the convolution operation. 

Starting from the input matrix $\mathbf{X}\in\mathbb{R}^{D \times n}$, the successive encoding mappings generate the layer-wise embeddings $\mathbf{Z}^{(\ell)}$, where $\mathbf{Z}^{(\ell)}$ is the embedding matrix of the $\ell^{\text{th}}$ layer. The output of the final layer \(L\) forms the latent representation \(\mathbf{Z} \in \mathbb{R}^{d \times n}\):

\begin{equation}
\begin{split}
    \mathbf{Z}^{(0)} &\;=\; \mathbf{X}, \\
    \mathbf{Z}^{(\ell)} &\;=\; f^{(\ell)}\!\bigl(\mathbf{Z}^{(\ell-1)}\bigr),\; \ell \in \{1,\dots,L \}\\
    \mathbf{Z} &\;=\; \mathbf{Z}^{(L)} \in \mathbb{R}^{d\times n},\qquad d\ll D.
\end{split}
\end{equation}

The decoding function $\mathscr{D}$ mirrors the encoder's architecture and reconstructs the input data \(\mathbf{X}\) from the latent representation \(\mathbf{Z}\). The reconstructed matrix from the latent code is denoted $\widehat{\mathbf{X}}$ and is expressed as follows:

\begin{equation}
    \widehat{\mathbf{X}} \;=\; \mathscr{D}\bigl(\mathbf{Z};\widehat{\mathbf{W}}\bigr)
                       \;=\; \mathscr{D}\bigl(\mathscr{E}(\mathbf{X};\mathbf{W});\widehat{\mathbf{W}}\bigr),
\end{equation}

\noindent where $\mathbf{W}$ and $\widehat{\mathbf{W}}$ constitute the collection of encoder and decoder parameters, respectively. 

The  convolutional auto-encoder is trained to minimize the mean-squared reconstruction loss:
\begin{equation}\label{eq:rec_loss}
    \mathcal{L}_{\text{rec}}
    \;=\; \frac{1}{n}\,\bigl\|\mathbf{X} - \widehat{\mathbf{X}}\bigr\|_{F}^{2}
    \;=\; \frac{1}{n}\,\bigl\|\mathbf{X} - \mathscr{D}(\mathscr{E}(\mathbf{X}))\bigr\|_{F}^{2},
\end{equation}
where $\|\cdot\|_{F}$ denotes the Frobenius norm.  
The compact representation $\mathbf{Z}$ obtained from the trained encoder feeds into the subsequent self-expressiveness module, which models pairwise relations in the latent space.

\subsection{Self-Expressiveness Property}

The self-expressiveness principle states that a point lying in a union of subspaces can be reconstructed as a linear combination of other points from the same subspace. A widely used sparse formulation is  \cite{paper6}, which is described as follows: 

\begin{equation}\label{eq:sparse}
    \min_{\mathbf{C}}\;\|\mathbf{C}\|_{1}\quad\text{s.t.}\quad \mathbf{X}=\mathbf{X}\mathbf{C}, \;\mathrm{diag}(\mathbf{C})=\mathbf{0},
\end{equation}

\noindent where \(\mathbf{C}\in\mathbb{R}^{n\times n}\) is the self-expressison matrix. The sparsity-promoting models \cite{paper6, paper17} rely on $l_{0}$ or $l_{1}$ penalties on \(\mathbf{C}\). Although these models yield interesting theoretical properties (e.g., subspace-preserving and block-diagonal self-expression matrix), the $l_{0}$ or $l_{1}$ terms can cause the over-segmentation problem \cite{paper18} and generally incur an expensive iterative optimization process \cite{paper17}. To address these problems, the least-squares regression (LSR) formulation for subspace clustering \cite{paper9} applies an $l_{2}$ penalty on $\mathbf{C}$. This particular variant permits the diagonal entries of $\mathbf{C}$ to be non-zero, while other formulations of LSR explicitly impose the constraint $\mathrm{diag}(\mathbf{C})=\mathbf{0}$. Allowing a non-zero diagonal leads to a denser subspace-preserving representation and admits a closed-form solution. The considered LSR optimization problem is as follows:

\begin{equation}
    \min_{\mathbf{C}}\; \frac12 \bigl\| \mathbf{X} - \mathbf{X}\mathbf{C} \bigr\|_{F}^{2} \;+\; \lambda \,\|\mathbf{C}\|_{F}^{2},
    \label{eq:LSR}
\end{equation}

\noindent where $\lambda$ is a balancing hyperparameter. The analytic solution of this problem is expressed as:

\begin{equation}
  \begin{aligned}
  \mathbf{C}^{\star} = (\mathbf{X}^{\top}\mathbf{X} + \lambda \, \mathbf{I}_{n})^{-1} \; \mathbf{X}^{\top}\mathbf{X}.
  \end{aligned}
\label{eq:analytic_solution}
\end{equation}

The principal objective is to reformulate the LSR problem so that its solution can be computed in linear time. A secondary objective is to embed the revised formulation within a deep clustering framework, thereby achieving linear-time performance even in the presence of random noise and outliers. Before detailing the scalable reformulation and the full integration, we highlight several key properties of the closed-form estimator \(\mathbf{C}^{\star}\) for the LSR subspace problem.

\begin{proposition}\label{proposition_1}
\textit{The matrix \(\mathbf{C}^{\star}\) is symmetric, positive semi-definite, and its rank is equal to the rank of \(\mathbf{X}\).}
\end{proposition}

These characteristics follow directly from the structure of the regularised Gram matrix \((\mathbf{X}^{\top}\mathbf{X} + \lambda \mathbf{I})^{-1}\mathbf{X}^{\top}\mathbf{X}\): symmetry and positive semi-definiteness are inherited from the inverse of the regularised Gram term, while rank preservation is a consequence of the Sylvester rank inequality. Detailed derivations are omitted since these are trivial properties.
\begin{definition} \label{defintion_1}
\textit{A square matrix $\mathbf{M} \in \mathbb{R}^{n \times n}$ is idempotent if and only if $\mathbf{M}^{2}=\mathbf{M}$.}
\end{definition}

\begin{proposition} \label{proposition_2}
\textit{Given the set $\left \{ \beta_i \right \}_{i=1}^{r}$ of the strictly positive eigenvalues of $\mathbf{X}^{\top}\mathbf{X}$ and the set $\left \{ \tau_i \right \}_{i=1}^{r}$ of the strictly positive eigenvalues of $\mathbf{C}^{\star}$, where $r$ is the rank of $\mathbf{C}^{\star}$, we assume that $\lambda$ is distributed uniformly on $\left [ 0, 1 \right ]$. Then: }

\begin{equation*}
\begin{split}
\forall \delta > 0,  \:\:\:\:  Pr\big(\left | \tau_{i} - 1  \right | \geqslant \delta \big) \leqslant \, \frac{1}{\delta} \: \left [1 +  \beta_{i} \: \text{ln}\Big(\frac{\beta_{i}}{1 + \beta_{i}}\Big) \right ].
\end{split}
\end{equation*}
\end{proposition}

\noindent  In Proposition \ref{proposition_2}, we demonstrate that the strictly positive eigenvalues of $\mathbf{C}^{\star}$ are nearly equal to $1$ with high probability when $\lambda \ll \beta_i, \; \forall i \in \left \{1, \cdots, r \right \}$. The probability bound is obtained by applying Markov's inequality to the random variable $1-\tau_i=\tfrac{\lambda}{\beta_i+\lambda}$ with $\lambda \sim \mathrm{Unif}[0,1]$. It is well-known that a matrix is idempotent if and only if all its eigenvalues are either $0$ or $1$. Based on Proposition \ref{proposition_2}, we can see that the matrix $\mathbf{C}^{\star}$ is nearly idempotent when $\lambda$ is fixed to a sufficiently small value (smaller than $\min_{i \in \left \{1, \cdots, r \right \}}(\beta_i)$). 

\subsection{Scalable Reformulation}

A possible idea is to approximate $\mathbf{C}^{\star}$ with the idempotent version of it. To reduce the $\mathcal{O}(n^{3})$ cost of solving problem~\eqref{eq:LSR} to linear time, we exploit the near-idempotence of $C^{\star}$ and show that it admits a rank-$r$ factorisation with high probability.

\begin{proposition} \label{proposition_3}
We assume that $\lambda$ is distributed uniformly on $\left [ 0, 1 \right ]$, then there exists a matrix $\mathbf{P} \in \mathbb{R}^{n \times r}$ such that:  
\begin{equation*}
\begin{cases}
\forall \delta > 0, \:  Pr\big(\left \| \mathbf{C}^{\star} - \mathbf{P} \mathbf{P}^{\top} \right \|_{F}^{2} \geqslant \delta \big) \leqslant \frac{1}{\delta} \: \sum_{i=1}^{r} \: g(\beta_{i}), \\
g(\beta_{i}) = \frac{2 \beta_{i} + 1}{\beta_{i} + 1} + 2 \: \beta_{i} \: \text{ln}(\frac{\beta_{i}}{\beta_{i} + 1}), \\
\mathbf{P}^{\top}\mathbf{P} = \mathbf{I}, \\
\operatorname{rank}(\mathbf{P})= \operatorname{rank}(\mathbf{C}^{\star}) = r.
\end{cases} 
\end{equation*}
\end{proposition}

\noindent Let $\mathbf{p}_{i}$ denote the eigenvectors of $\mathbf{C}^{\star}$ associated with the strictly positive eigenvalues $\tau_{i}$ and set $\mathbf{P}=[\mathbf{p}_{1},\dots,\mathbf{p}_{r}]\in\mathbb{R}^{n\times r}$.  Proposition~\ref{proposition_3} states that $\mathbf{C}^{\star}$ can be approximated by the projector $\mathbf{P}\mathbf{P}^{\top}$ with high probability, bounded by $\frac{1}{\delta}\sum_{i=1}^{r}\!\bigl[\frac{2\beta_{i}+1}{\beta_{i}+1}+2\beta_{i}\ln\bigl(\frac{\beta_{i}}{\beta_{i}+1}\bigr)\bigr]$, where $\beta_{i}$ are the positive eigenvalues of $\mathbf{X}^{\top}\mathbf{X}$. The inequality in this proposition is likewise derived via Markov's inequality, applied to $\|C^\star - PP^\top\|_F^2 = \sum_{i=1}^r \bigl(\tfrac{\lambda}{\beta_i+\lambda}\bigr)^2$ with $\lambda \sim \mathrm{Unif}[0,1]$. To verify the tightness of this bound, we generated ten i.i.d.\ Gaussian matrices $\mathbf{X}\!\sim\!\mathcal{N}(0,1)^{\,r\times 1000}$ for each rank $r\in\{1,\dots,100\}$, fixed $\delta=10^{-3}$, and computed both the bound and the empirical probability of $\|\mathbf{C}^{\star}-\mathbf{P}\mathbf{P}^{\top}\|_{F}^{2}\le\delta$.  Across all $1{,}000$ trials the empirical probability exceeded $0.9992$, confirming that $\mathbf{C}^{\star}$ is accurately approximated by $\mathbf{P}\mathbf{P}^{\top}$ with high probability.

According to Proposition \ref{proposition_3}, we can reformulate the optimization problem in Eq. (\ref{eq:LSR}) as follows:

\begin{equation}
\begin{aligned}
&\min_{\mathbf{P}\in\mathbb{R}^{n\times r}}
   \;\bigl\|\mathbf{X}-\mathbf{X}\mathbf{P}\mathbf{P}^{\top}\bigr\|_{F}^{2}
   +\lambda \, \|\mathbf{P}\mathbf{P}^{\top}\|_{F}^{2}\\
&\text{s.t. }\;
   \mathbf{P}^{\top}\mathbf{P}=\mathbf{I}_{r},\;
   \operatorname{rank}(\mathbf{X}) = \operatorname{rank}(\mathbf{P})=r.
\end{aligned}
\label{eq:LRSC}
\end{equation}

Since the columns of \(\mathbf{P}\) are orthonormal, \(\|\mathbf{P}\mathbf{P}^{\top}\|_{F}^{2}=\|\mathbf{P}\|_{F}^{2}= 
\operatorname{tr}(\mathbf{P}^{\top}\mathbf{P}) = r\). Therefore, the regularization term is a constant and can be dropped. The problem in Eq. (8) can be solved in linear time if we know the rank of the matrix $\mathbf{X}$. More precisely, we can expand the squared Frobenius norm:

\begin{equation}
\bigl\|\mathbf{X}-\mathbf{X}\mathbf{P}\mathbf{P}^{\top}\bigr\|_{F}^{2}
      = \|\mathbf{X}\|_{F}^{2}-\operatorname{tr}
        \bigl(\mathbf{P}^{\top}\mathbf{X}^{\top}\mathbf{X}\mathbf{P}\bigr).
\label{eq:norm}
\end{equation}

Assuming that we know the rank of $\mathbf{X}$, minimizing the objective in Eq. (\ref{eq:LRSC}) is thus equivalent to solving the problem:

\begin{equation}
\max_{\mathbf{P}^{\top}\mathbf{P}=\mathbf{I}_{r}}\;
\operatorname{tr}\bigl(\mathbf{P}^{\top}\mathbf{X}^{\top}\mathbf{X}\mathbf{P}\bigr).
\label{eq:LRSC_without_rank}
\end{equation}

The problem in Eq.~\eqref{eq:LRSC_without_rank} is equivalent to the classical rank–\(r\) PCA problem.  By the Rayleigh–Ritz theorem, the maximum is obtained by choosing the columns of \(\mathbf{P}\) as the \(r\) eigenvectors of
\(\mathbf{X}^{\top}\mathbf{X}\) corresponding to its largest eigenvalues.
Let \(\mathbf{X}= \mathbf{U}\boldsymbol{\Sigma}\mathbf{V}^{\top}\) be a truncated SVD, then \(\mathbf{P}= \mathbf{V}_{(:,1:r)}\) is the global optimum and
\(\mathbf{P}\mathbf{P}^{\top}=\mathbf{V}_{(:,1:r)}\mathbf{V}_{(:,1:r)}^{\top}\) is the rank-\(r\) projector that minimises the cost function in Eq~\eqref{eq:LRSC}.

The optimal $\mathbf{P}$ is given by the $r$ dominant right singular vectors of $\mathbf{X}$, obtainable via a truncated SVD in $\mathcal{O}(n)$ assuming that we know the rank of the matrix $\mathbf{X}$. Nevertheless, the rank is not provided, and estimating it requires a quadratic computational complexity. To address this issue, we reformulate the problem in Eq.~\eqref{eq:LRSC} by substituting the matrix $\mathbf{X} \, \mathbf{P}$ with a landmark matrix $\mathbf{L} \in \mathbb{R}^{D \times m}$, where $m$ is the number of anchor points and $m\ll n$. Thus, we obtain the problem:

\begin{equation}
\min_{\mathbf{P}\in\mathbb{R}^{n\times m}, \:\:\mathbf{L}\in\mathbb{R}^{D\times m}} \;\bigl\|\mathbf{X}-\mathbf{L} \, \mathbf{P}^{\top} \bigr\|_{F}^{2} \quad\text{s.t. }\;
   \mathbf{P}^{\top}\mathbf{P}=\mathbf{I}_{m}.
\label{eq:landmark}
\end{equation}
The rows of $\mathbf{L}$ constitute the anchor points and serve as a representative subset of the data, while the orthogonal $\mathbf{P}$ supplies the assignment weights. 

Embedding the input matrix $\mathbf{X}$ in the latent space of the convolutional auto-encoder, we jointly minimize reconstruction and subspace terms:

\begin{equation}
\begin{aligned}
\min_{\mathbf{W},\,\widehat{\mathbf{W}},\,\mathbf{P},\,\mathbf{L}}\;
        &\frac{1}{n}\,\bigl\|\mathbf{X}-\widehat{\mathbf{X}}\bigr\|_{F}^{2}
         +\bigl\|\mathbf{Z}-\mathbf{L} \, \mathbf{P}^{\top}\bigr\|_{F}^{2}\\
\text{s.t.}\;\;\;&\mathbf{P}^{\top}\mathbf{P}=\mathbf{I}_{m},
\end{aligned}
\label{eq:joint_obj}
\end{equation}

\subsection{Optimization} \label{sec:optimization}

We minimise the joint objective in Eq.~\eqref{eq:joint_obj} with a \emph{block-coordinate descent} scheme that cyclically updates three disjoint variable blocks: $(\mathbf{W},\widehat{\mathbf{W}}),\, \mathbf{P},\,\mathbf{L}$.

At every step, all but one block are held fixed. Thus, the global problem is turned into tractable subproblems that admit either a closed-form or an efficient gradient update. 

\vspace{4pt}
\noindent\textbf{(i) Network parameters $\mathbf{W},\widehat{\mathbf{W}}$.} With $\mathbf{P}$ and $\mathbf{L}$ frozen, Eq.~\eqref{eq:joint_obj} reduces
to the sum of a reconstruction loss and a quadratic subspace term that is differentiable w.r.t. the auto-encoder weights. We therefore apply
Adam with back-propagation to update the network parameters. The per-epoch cost for updating  $\mathbf{W}$ and $\widehat{\mathbf{W}}$ is $\mathcal{O}(nd)$, assuming all layers have a dimension equal to or lower than $d$.

\vspace{4pt}
\noindent\textbf{(ii) Orthogonal projector $\mathbf{P}$.}\label{par:p_update} Fixing $\mathbf{W},\widehat{\mathbf{W}},\mathbf{L}$, we obtain the classical \emph{orthogonal Procrustes} problem:

\begin{equation}
\min_{\mathbf{P}^{\top}\mathbf{P}=I_{m}}\;\|\mathbf{Z}-\mathbf{L} \, \mathbf{P}^{\top}  \|_{F}^{2}.
\label{eq:procrustes}
\end{equation}

This problem has a closed-form solution obtained by computing the SVD of $\mathbf{Z}^{\top}\mathbf{L}= \mathbf{U}' \mathbf{\Sigma}' \mathbf{V}'^{\top}$. The global minimizer is \(\mathbf{P}^{\star}= \mathbf{U}'\mathbf{V}'^{\top}\). 
Forming \(\mathbf{Z}^{\top}\mathbf{L}\) costs \(\mathcal{O}(ndm)\) and the SVD of the resulting \(n\times m\;(n\!\gg\!m)\) matrix costs a further \(\mathcal{O}(nm^{2})\), so the entire update is linear in the sample size \(n\).

\vspace{4pt}
\noindent\textbf{(iii) Landmark matrix $\mathbf{L}$.}\ With $\mathbf{W},\widehat{\mathbf{W}},\mathbf{P}$ fixed, we obtain the least-squares problem:

\begin{equation}
\min_{\mathbf{L}}\;\|\mathbf{Z}-\mathbf{L} \,  \mathbf{P}^{\top}  \|_{F}^{2}.
\label{eq:least_square_solution}
\end{equation}

This problem has a closed-form solution $\mathbf{L}^{\star}= \mathbf{Z} \, \mathbf{P}$, which costs $\mathcal{O}(nmd)$ operations. Since Eq. (\ref{eq:least_square_solution}) is a convex problem, it can also be solved using gradient descent. 

\vspace{4pt}
\noindent\textbf{Complexity per epoch.}\
One full cycle over the three blocks therefore costs \(\mathcal{O}(n\,d)+\mathcal{O}(nm^{2})+\mathcal{O}(nmd)\), which is linear in \(n\) for fixed $m$ and $d$.

\vspace{4pt}
\noindent\textbf{Summary.}\
The optimization procedure is listed in Algorithm 1. We initialise \(\mathbf{W},\widehat{\mathbf{W}}\) with a standard reconstruction using the convolutional auto-encoder as a pretraining phase, choose $m$ anchor points via $k$-means++\cite{paper47}, set \(\mathbf{L}\) to those anchors, and obtain an initial \(\mathbf{P}\) with the Procrustes step. The alternating updates typically converge in fewer than ten outer iterations. In all our experiments, the objective decreased monotonically, and no numerical instability was observed.

\subsection{Clustering Step}
After learning the latent representation and the decomposition of the self-expressiveness matrix, the final step is to cluster the data into \(k\) groups. This is achieved through a scalable spectral clustering approach. Usually, spectral clustering relies on the eigendecomposition of the graph Laplacian derived from the affinity matrix \(\mathbf{S}\). However, constructing and decomposing the full affinity matrix \(\mathbf{S} \in \mathbb{R}^{n \times n}\) is computationally expensive, with a complexity of \(\mathcal{O}(n^3)\). To address this, we propose a scalable approach that leverages the matrix \(P \in \mathbb{R}^{n \times m}\) to reduce the computational burden.

\vspace{3pt}
\noindent\textbf{Affinity in factored form.} From Section~\ref{sec:optimization} we already have the decomposition $\mathbf{C}=\mathbf{P}\mathbf{P}^{\top}$ with $\mathbf{P}\!\in\!\mathbb{R}^{n\times m}$ and $m\!\ll\!n$, which immediately ensures a symmetric, positive-semidefinite affinity matrix $\mathbf{C}\!\in\!\mathbb{R}^{n\times n}$. Because these properties are guaranteed by construction, we can feed $\mathbf{C}$ directly into the spectral-clustering pipeline without the additional symmetrisation step $\mathbf{S}=\tfrac{|\mathbf{C}|+|\mathbf{C}^{\top}|}{2}$ often used in earlier work.

\vspace{3pt}
\noindent\textbf{Graph Laplacian via SVD.} The graph Laplacian $\mathbf{\mathscr{L}}$ is computed from the affinity matrix \(\mathbf{S}\). In our case, $\mathbf{S} = \mathbf{C}$. Then, we adopt the \emph{unnormalised} graph Laplacian defined as:
\begin{equation}\label{graph_laplacian}
       \mathbf{\mathscr{L}}=\mathbf{D}-\mathbf{C},
\end{equation}

\noindent where \(\mathbf{D}\) is the degree matrix of the affinity graph, which is a diagonal matrix with \(\mathbf{D}_{ii} = \sum_{j=1}^n \mathbf{C}_{ij}\). Let the SVD of $\mathbf{P}$ be $\mathbf{P}= \mathbf{Q}\boldsymbol{\Lambda}\mathbf{R}^{\top}$, such that $\mathbf{Q}\in\mathbb{R}^{n\times m}$, $\boldsymbol{\Lambda}\in\mathbb{R}^{m\times m}$, an $\boldsymbol{R}\in\mathbb{R}^{m\times m}$. Thus, we can write   
$\mathbf{C}= \mathbf{Q}\boldsymbol{\Lambda}^{2}\mathbf{Q}^{\top}$. Because \(\mathbf{\mathscr{L}}\) and \(\mathbf{C}\) share the column
space of \(\mathbf{Q}\), the eigenvectors of \(\mathbf{\mathscr{L}}\) with the \(k\) smallest eigenvalues are
obtained by solving the much smaller \(m\times m\) problem  
\(
\bigl(\mathbf{Q}^{\top}\mathbf{D}\mathbf{Q}-\boldsymbol{\Lambda}^{2}\bigr)\mathbf{y}
   =\mu\,\mathbf{y}
\)
and lifting the solutions via \(\mathbf{h}=\mathbf{Q}\mathbf{y}\). The cost is dominated by the initial SVD of \(\mathbf{P}\), \(\mathcal{O}(nm^{2})\), instead of the \(\mathcal{O}(n^{3})\) required for a full eigendecomposition of the normalized graph Laplacian.

\vspace{3pt}
\noindent\textbf{\(k\)-means in embedding space.}  
Stack the \(k\) lifted eigenvectors \(\mathbf{h}_{1},\dots,\mathbf{h}_{k}\) into the matrix \(\mathbf{H}_{k}\in\mathbb{R}^{n\times k}\). Each row of \(\mathbf{H}_{k}\) is the spectral embedding of a sample under the unnormalised Laplacian. Apply \(k\)-means to these rows.  The Lloyd iterations cost \(\mathcal{O}(nk^{2}t)\), which is negligible relative to the preceding \(\mathcal{O}(nm^{2})\) SVD because \(k\!\ll\!m\).

\vspace{3pt}
\noindent\textbf{Complexity.}  
The dominant cost of the clustering stage is the SVD computation of \(\mathbf{P}\), which runs in \(\mathcal{O}(nm^{2})\) time. Solving the \(m\times m\) eigenproblem adds \(\mathcal{O}(m^{3})\) computation complexity, and \(k\)-means contributes \(\mathcal{O}(nk^{2}t)\) for \(t\) Lloyd iterations. Because \(m,k\!\ll\!n\), the \(nm^{2}\) term governs the runtime, while the working memory is \(\mathcal{O}(nm)\) for the factor matrix and its singular vectors. Hence, the entire clustering step remains linear in the sample size \(n\) for a fixed anchor count \(m\). A concise overview of the full pipeline is presented in Algorithm~\ref{alg:scaldeep}.

\begin{algorithm}[t]
\caption{\textsc{SDSNet}}
\label{alg:scaldeep}
\KwIn{data matrix $\mathbf{X}$, target clusters $k$, anchors $m\;(m\!\ll\!n)$, regulariser $\lambda$, CAE depth $L$, latent size $d$.}
\KwOut{cluster labels $\{1,\dots,k\}$ for all samples.}

\textbf{1.~Feature extraction}\;
\Indp
\textit{Pre-train CAE:}  minimise the reconstruction loss
$\mathcal{L}_{\text{rec}}$ in~\eqref{eq:rec_loss} and obtain embeddings
$\mathbf{Z}=\mathscr{E}(\mathbf{X};\mathbf{W})\in\mathbb{R}^{d\times n}$.\;
\Indm

\textbf{2.~Anchor initialisation}\;
\Indp
Choose $m$ anchors by $k$-means++ in~$\mathbf{Z}$;
let $\mathbf{L}^{(0)}\in\mathbb{R}^{d\times m}$ be the anchor matrix
(the columns of $\mathbf{L}^{(0)}$ is a subset of the columns of $\mathbf{Z}$) and obtain
$\mathbf{P}^{(0)}$ from one Procrustes step
\eqref{eq:procrustes}.\;
\Indm

\textbf{3.~Joint optimisation (block coordinate descent)}\;
\Indp
\While{not converged}{
    \textit{(a) Update CAE weights} $(\mathbf{W},\widehat{\mathbf{W}})$
    by one Adam step on the joint loss~\eqref{eq:joint_obj}\;
    \textit{(b) Update $\mathbf{P}$} :
    SVD of $\mathbf{Z}^{\top}\mathbf{L}$,
    $\mathbf{Z}^{\top}\mathbf{L}=\mathbf{U}'\boldsymbol{\Sigma}'\mathbf{V}'^{\top}$,
    then $\mathbf{P}\leftarrow\mathbf{U}'\mathbf{V}'^{\top}$\;
    \textit{(c) Update $\mathbf{L}$} : closed-form
    $\mathbf{L}\leftarrow\mathbf{Z}\mathbf{P}$\;
}
\Indm

\textbf{4.~Spectral embedding without forming \(\mathbf{C}\)}\;
\Indp
\begin{enumerate}
\item \emph{Compute SVD}\,: $\mathbf{P}=\mathbf{Q}\boldsymbol{\Lambda}\mathbf{R}^{\top}$
      with \(\mathbf{Q}\!\in\!\mathbb{R}^{n\times m}\);
\item \emph{Compute degree vector in \(O(nm)\)}\,:\\[-4pt]
      \begin{align*}
      \mathbf{s} &\leftarrow \mathbf{P}^{\top}\mathbf{1}_{n};\quad &&\text{//\(m\) inner products}\\
      \mathbf{d} &\leftarrow \mathbf{P}\mathbf{s};\quad            &&\text{//matrix–vector product}
      \end{align*}
\item \emph{Reduced eigenproblem}\,: 
      \(\mathbf{M}= \mathbf{Q}^{\top}\operatorname{diag}(\mathbf{d})\mathbf{Q}-\boldsymbol{\Lambda}^{2}\) \\
      Solve \(\mathbf{M}\mathbf{y}=\mu\mathbf{y}\) for the
      \(k\) smallest eigenpairs; \\
      Compute \(\mathbf{H}_{k}= \mathbf{Q}\mathbf{Y}_{k}\);
\end{enumerate}
\Indm

\textbf{5.~Clustering}\;
\Indp
Run $k$-means on the rows of $\mathbf{H}_{k}$; \\
Return clustering assignments\;
\Indm
\end{algorithm}

\begin{table}[t]
  \begin{center}
   \caption{Description of real-world datasets.}
   \label{Table:description_data}
  \begin{footnotesize}
  
  \begin{tabular}{|c|c|c|c|}
    \hline
   {\textbf{Dataset}} & {\textbf{Size}} & 
    {\textbf{Groups}} & {\textbf{Dimensions}} 
     \\ \hline
   
    \textbf{YaleB} & 2,432 & 38 & 48 $\times$ 42  \\ \hline
    \textbf{ORL} & 400 & 40 & 32 $\times$ 32  \\ \hline
    \textbf{Coil100} & 7,200 & 100 & 32 $\times$ 32\\ \hline
    \textbf{UMIST} & 480 & 20 & 32 $\times$ 32\\ \hline
    \textbf{Fashion-MNIST} & 60,000 & 10 & 28 $\times$ 28\\ \hline

  \end{tabular}
  \end{footnotesize}
  \end{center}
  \vskip -0.1in
  
\end{table}

\begin{table*}[t]
 
  \begin{center}
   \caption{Comparison with scalable subspace clustering approaches. Bold indicates the best result in each column. ACC, NMI, and SPE are reported as percentages (\%). M indicates memory limit.}
   \label{Quantitative-study1}
  \resizebox{\linewidth}{!}{
  \begin{tabular}{|c|c|c|c|c|c|c|c|c|c|c|c|c|c|c|c|}
    \hline
   {\textbf{Scalable Methods}} &  \multicolumn{3}{c|}{\textbf{Yaleb }} &  \multicolumn{3}{c|}{\textbf{UMIST }} &\multicolumn{3}{c|}{\textbf{ORL }}&
     \multicolumn{3}{c|}{\textbf{Coil100 }}& \multicolumn{3}{c|}{\textbf{Fashion-Mnist }} \\
    \cline{1-16}
    
    \textbf{Shallow }  & \textbf{ACC}$\uparrow$ & \textbf{NMI}$\uparrow$ & \textbf{SPE}$\downarrow$ & \textbf{ACC}$\uparrow$ & \textbf{NMI}$\uparrow$ & \textbf{SPE} $\downarrow$ & \textbf{ACC}$\uparrow$ & \textbf{NMI}$\uparrow$ & \textbf{SPE}$\downarrow$ & \textbf{ACC}$\uparrow$ & \textbf{NMI}$\uparrow$ & \textbf{SPE}$\downarrow$ & \textbf{ACC}$\uparrow$ & \textbf{NMI}$\uparrow$ & \textbf{SPE}$\downarrow$ \\ \hline
    \hline
   \textbf{SSC-OMP}  &  75.03 & 80.27 & 0.17 &  67.7 & 77.7 & 0.30 &  70 & 84.44 & 0.28 & 59.46 & 82.02 & 0.28 &  36.24 & 42.78 & 0.57 \\ \hline
   \textbf{EnSC}  &  88.19 & 89.02 & 0.30 &  61.03 & 74.12 & 0.35  &  79.43 & 90.28 & 0.63 & 67.34 & 88.53 & 0.25 &  61.35 & 62.27 & 0.28 \\ \hline
    \textbf{SSSC}  &  80.67 & 84.78 & 0.19  &  61.79 & 72.95 & 0.45  &  74.72 & 85.43 & 0.52 & 66.89 & 88.11 & \textbf{0.03}  &  60.50 & \textbf{63.12} & \textbf{0.23} \\ \hline
   \textbf{A-DSSC}  &  91.7 & 94.7 & \textbf{0.08} &  72.5 & 85.1 & \textbf{0.03} &  79.0 & \textbf{91.0} & \textbf{0.15} & \textbf{82.4} & \textbf{94.6} & \textbf{0.03} & - & - & -  \\ \hline
    \textbf{LMVSC}  &  21.5 & 45.74 & 0.32  &  60.62 & 74.80 & 0.4 &  50 & 68.24 & 0.49 & 49.5 & 71.78 & 0.55 &  59.18 & 50.44 & M \\ \hline
   \textbf{SGL}  &  21.5 & 47.58 & 0.78 &  65 & 78.48 & 0.16  &  56.75 & 70.80 & 0.58 & 47.57 & 72.33 & 0.66 &  54.40 & 55.95 & M\\ \hline
    \textbf{S$^5$C}  &  60.81 & 65.55 & 0.29 &  69.79 & 78.62 & 0.08  &  72.75 & 85.21 & 0.44 & 55.25 & 79.70 & 0.17 &  59.78 & 61.74 & M\\ \hline
    \hline
   \textbf{SDSNet}  & \textbf{ 96.46} & \textbf{95.49} & 0.24  & \textbf{ 83.33} & \textbf{90.15} & 0.09  & \textbf{ 85.5} & 89.60 & 0.54  & 71.10 & 90.95 & 0.24 & \textbf{ 61.86} & 61.39 & 0.61\\ \hline
    
  \end{tabular}
}
  \end{center}
  \vskip -0.1in
 
\end{table*}

\begin{table}[t]
  \begin{center}
   \caption{Comparison with non-scalable subspace clustering approaches on ORL, Yaleb, and Coil100.}
   \label{Quantitative-study2}
  \begin{small}
  
  \begin{tabular}{|c|c|c|c|}
    \hline
   {\textbf{Non-Scalable Methods}} & 
    {\textbf{ORL}} & {\textbf{Yaleb}} & {\textbf{Coil100}}\\ \hline
    \textbf{ Deep} & \textbf{ACC} & \textbf{ACC} & \textbf{ACC} \\ \hline
    \hline
    \textbf{ DSC} & - & - & 69.63 \\ \hline
    \textbf{ DSCNet} & 86 &  97.33 & 69.04\\ \hline
    \textbf{DLRSC}& 83 & 97.53 & 71.86\\ \hline
    \textbf{S$^2$ConvSCN} & \textbf{89.5} & \textbf{98.48} & \textbf{73.33}\\ \hline
    \hline
    \textbf{SDSNet} & 85.5 & 96.46 & 71.10 \\ \hline
  \end{tabular}
  \end{small}
  \end{center}
  \vspace{-4mm}
\end{table}

\section{Experiments}

\subsection{Experimental Setup}

To evaluate the performance of SDSNet, we compare it with seven state-of-the-art scalable subspace clustering methods and four deep subspace clustering methods. The scalable subspace clustering approaches include methods based on sampling techniques, namely SGL \cite{paper36}, S$^5$C \cite{paper38}, and LMVSC \cite{paper15}, as well as methods that do not rely on sampling, namely SSC-OMP \cite{paper17}, EnSC \cite{paper19}, SSSC \cite{paper18}, and A-DSSC \cite{paper22}. For deep subspace clustering, we include DSC \cite{paper43}, DSCNet \cite{paper28}, S$^2$ConvSCN \cite{paper32}, and DLRSC \cite{paper39}. While these methods are not scalable, they serve as benchmarks for clustering accuracy. We evaluate the performance of competing algorithms on five widely used real-world datasets. Table \ref{Table:description_data} provides a summary description of these datasets.


To assess the clustering results, we use four evaluation metrics: Accuracy (ACC), Normalized Mutual Information (NMI), Subspace Preserving Error (SPE)~\cite{paper7}, and Connectivity (CONN)~\cite{paper18}. ACC and NMI both take values in $[0,1]$, with higher values indicating better clustering quality. SPE also lies in $[0,1]$, but lower values are better since they indicate fewer violations of the subspace-preserving property. As for CONN, it is defined as the second smallest eigenvalue of the normalized Laplacian, with larger values corresponding to better graph connectivity.


For SGL, S$^5$C, LMVSC, SSC-OMP, EnSC, and SSSC, we use publicly available source codes provided by the authors and fine-tune the parameters for optimal performance on each dataset. Since the source code for A-DSSC is unavailable, we rely on the results published in its original paper for the YaleB, ORL, and UMIST datasets. However, A-DSSC was not evaluated on Fashion-MNIST in the original work, and thus, no results are reported for it on that dataset. For deep subspace clustering models (DSC, DSCNet, S$^2$ConvSCN, and DLRSC), we report results from previously published research articles. Notably, these experiments were conducted only on YaleB, ORL, and Coil100, and the NMI, SPE, and CONN metrics were not evaluated in those studies. Consequently, for these datasets, we compare only the clustering accuracy (ACC) as reported in the original papers.

\begin{figure*}

  \begin{center}
    \begin{subfigure}[b]{0.3\textwidth}
    \includegraphics[width=0.7\linewidth]{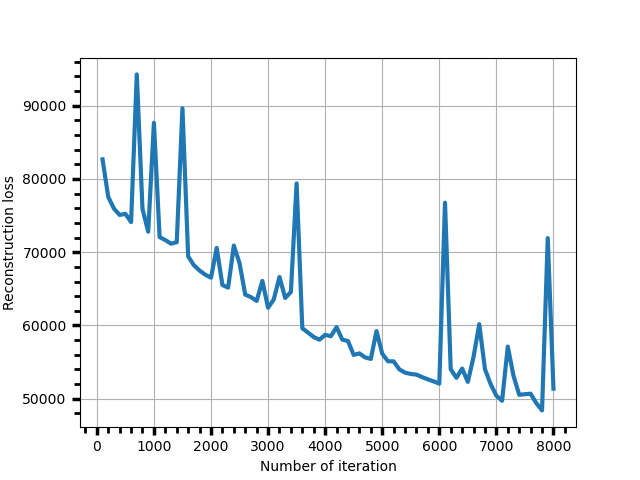}
    \caption{Training phase.}
   \end{subfigure}
  \begin{subfigure}[b]{0.3\textwidth}
     \includegraphics[width=0.7\linewidth]{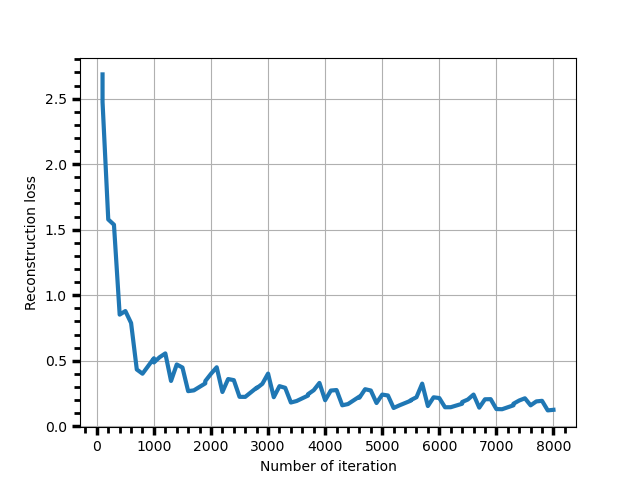}
     \caption{Pre-training phase.}
   \end{subfigure}
 \begin{subfigure}[b]{0.3\textwidth}
     \includegraphics[width=0.7\linewidth]{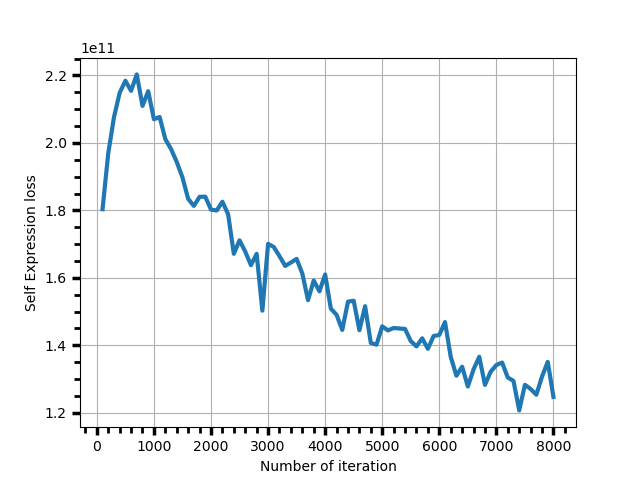}
     \caption{Self-expression loss function.}
   \end{subfigure}
   \vskip 0.1in
  \caption{The evolution of different loss functions outer iterations on Yaleb data base.}
  \label{cost-training}
  \end{center}
  
\end{figure*}

\begin{figure*}
  \begin{center}
    \begin{subfigure}[b]{0.33\textwidth}
    \includegraphics[width=0.75\linewidth]{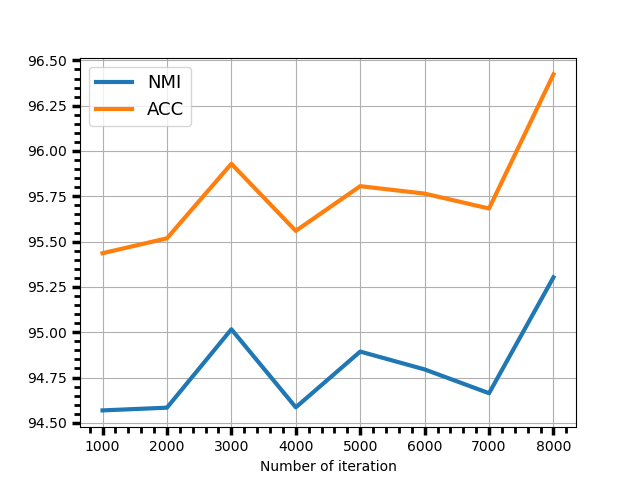}

  \end{subfigure}
 \begin{subfigure}[b]{0.33\textwidth}
     \includegraphics[width=0.75\linewidth]{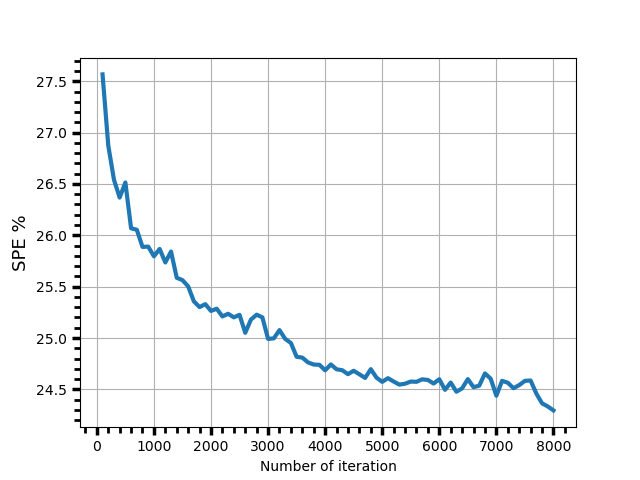}

  \end{subfigure}
   \vskip 0.1in
  \caption{The evolution of ACC, NMI, and SPE outer iterations on Yaleb database.}
  \label{evolution}
  \end{center}

\end{figure*}

\subsection{Clustering Quality}

Table \ref{Quantitative-study1} presents the clustering performance of our approach compared to scalable subspace clustering methods. Overall, the results indicate that our method outperforms LMVSC, SGL, and S$^5$C, which rely on sampling techniques. Specifically, our approach surpasses LMVSC and SGL by an average of 74.96\% for YaleB, 20.52\% for UMIST, 32.12\% for ORL, 22.56\% for Coil100, and 5.36\% for Fashion-MNIST. Unlike our method, LMVSC and SGL depend on random selection, which is ineffective in identifying subspaces. Although S$^5$C improves subspace identification over LMVSC and SGL by employing a targeted selection strategy, it is constrained by overly restrictive theoretical conditions and remains less effective. Our approach outperforms S$^5$C by an average of 35.65\% for YaleB, 13.54\% for UMIST, 12.75\% for ORL, 15.85\% for Coil100, and 2.66\% for Fashion-MNIST.

Furthermore, our method achieves higher clustering performance than SSC-OMP, EnSC, SSSC, and A-DSSC, with an average improvement of 12.56\% for YaleB, 17.57\% for UMIST, 9.71\% for ORL, 2.07\% for Coil100, and 10.28\% for Fashion-MNIST. Unlike traditional methods, our approach leverages deep learning to effectively learn a subset of data that enhances subspace identification while maintaining computational efficiency. Furthermore, our results on NMI and SPE confirm that our approach effectively preserves subspace structures.

Table \ref{Quantitative-study2} compares the clustering accuracy of our approach with deep subspace clustering methods that are not scalable. These methods construct the affinity matrix using the similarity between each point and all other data points. Our results are competitive with those obtained by deep subspace clustering approaches, with a maximum difference of 4\% for ORL, 2.02\% for YaleB, and 2.23\% for Coil100.

\

\begin{figure*}
  \begin{center}
   
  \begin{subfigure}[b]{0.33\textwidth}
    \includegraphics[width=0.85\linewidth]{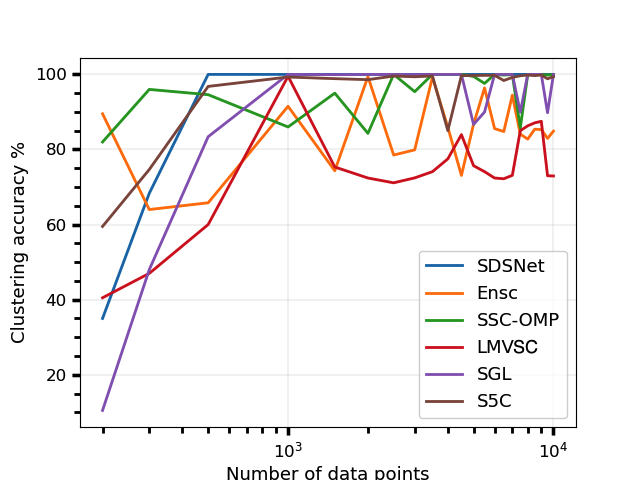}
    
      \end{subfigure}
    \begin{subfigure}[b]{0.33\textwidth}
     \includegraphics[width=0.85\linewidth]{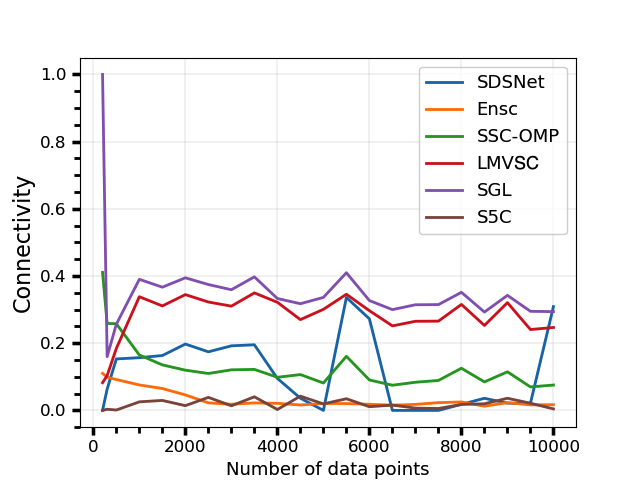}
  
      \end{subfigure}\\
    \begin{subfigure}[b]{0.33\textwidth}
    \includegraphics[width=0.85\linewidth]{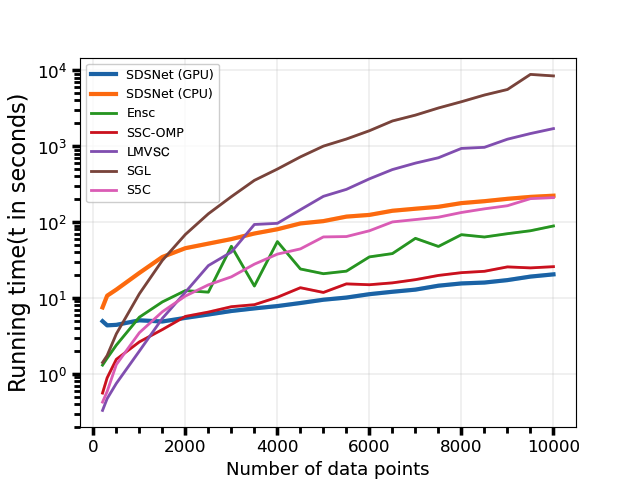}

  \end{subfigure}
    \begin{subfigure}[b]{0.33\textwidth}
    \includegraphics[width=0.85\linewidth]{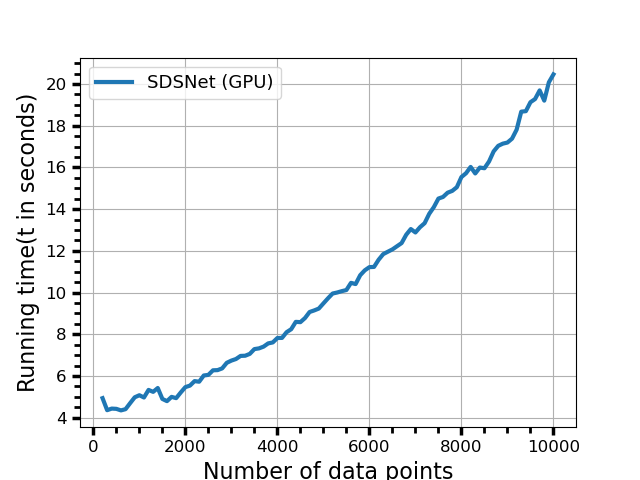}
  
    \end{subfigure}
     \vskip 0.1in
  \caption{Performance comparison of EnSC, SSC-OMP, LMVSC, SGL, S5C, and SDSNet on Synthetic data.}
   \label{Scalability}
  \end{center}
  
\end{figure*}

\begin{figure*}[ht]
  \begin{center}
    \begin{subfigure}[b]{0.33\textwidth}
    \includegraphics[width=\linewidth]{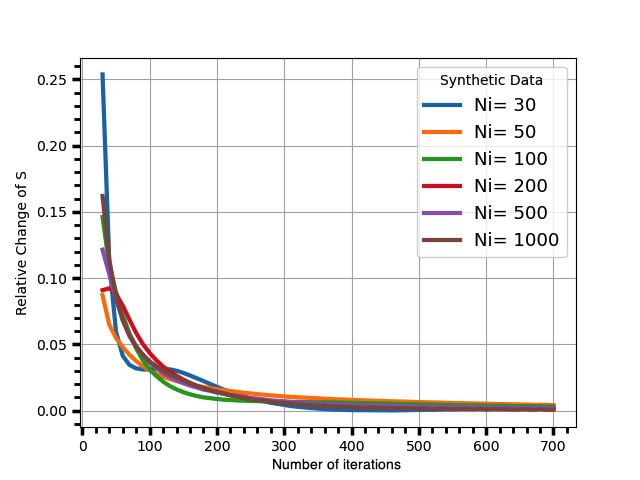}
    
  \end{subfigure}
  \begin{subfigure}[b]{0.33\textwidth}
     \includegraphics[width=\linewidth]{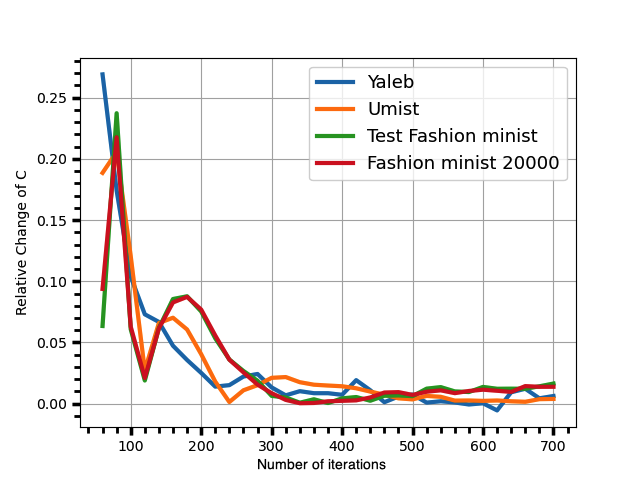}
    
  \end{subfigure}
 
   \vskip 0.1in
  \caption{The relative changes of S in successive outer iterations. \textit{Ni: \#data points in each subspace, \# of subspace = 10}.}
  \label{relative change}
  \end{center}
\end{figure*}

In addition, we analyze the evolution of different cost functions, including ACC, NMI, and SPE, across iterations. Figures \ref{cost-training} and \ref{evolution} illustrate the experimental results on the YaleB dataset. As observed, the reconstruction loss and self-representation loss steadily decrease over iterations while clustering accuracy and NMI improve, and the subspace-preserving error decreases.

\subsection{Scalability Analysis}

In this section, we study the scalability of our approach with respect to the dataset size. To this end, we generate synthetic datasets with an increasing number of data points. The generation process follows the configuration used in \cite{paper18}, generating $s = 10$ subspaces with random dimensions $d_i \in [6, 12]$ in an ambient space of $D = 784$. Each subspace contains $n_i$ randomly generated data points, where $n_i$ varies from 30 to 1,000. This results in total data points $n$ ranging from 300 to 10,000. Figure \ref{Scalability} presents a comparison of results on clustering accuracy, connectivity, and execution time for constructing the affinity matrix. Since the selected scalable subspace clustering methods for comparison run on a CPU, we present two execution time results for SDSNet: (1) execution time with a GPU and (2) execution time with a CPU.

We observe that SDSNet achieves the best clustering accuracy while maintaining the lowest computational cost with a GPU. When using a CPU, the computational cost of SDSNet is comparable to or even lower than that of other methods. Additionally, SGL produces better connectivity due to its use of a connectivity constraint to generate a bipartite graph with $k$ connected components. However, SGL incurs the highest computational cost among the compared methods.

\subsection{Convergence of the Affinity Matrix}

To evaluate convergence, we analyze the relative change of the affinity matrix $C$ between two successive iterations. Figure \ref{relative change} presents the results on synthetic and real datasets. We observe that the affinity matrix stabilizes after a few iterations, confirming the convergence of our algorithm.

\begin{figure*}
  \begin{center}
    \begin{subfigure}[b]{0.3\textwidth}
    \includegraphics[width=0.75\linewidth]{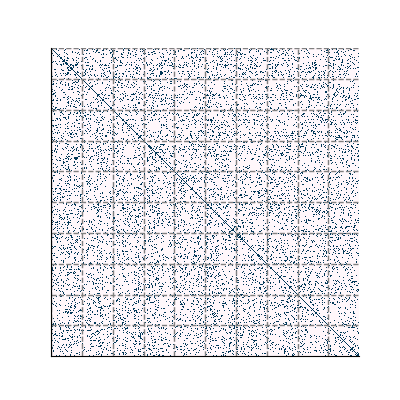}
    \caption{Epoch 0.}
    
  \end{subfigure}
  \begin{subfigure}[b]{0.3\textwidth}
     \includegraphics[width=0.75\linewidth]{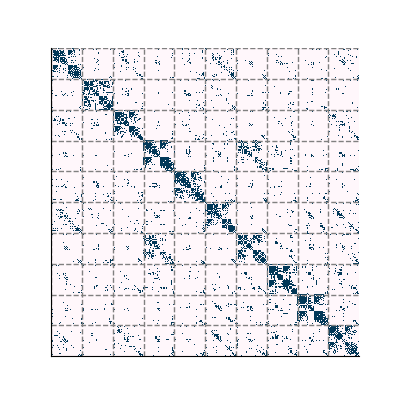}
     \caption{Epoch 100.}
  
  \end{subfigure}
 \begin{subfigure}[b]{0.3\textwidth}
     \includegraphics[width=0.75\linewidth]{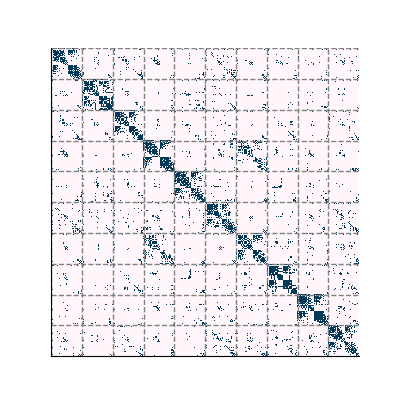}
     \caption{Epoch 500.}
   
  \end{subfigure}
  \caption{Visualizing the graph formed by the affinity matrix on Yaleb with 10 classes.}
  \label{graph YaleB}
  \end{center}
  \vspace{-4mm}
\end{figure*}

Furthermore, Figure \ref{graph YaleB} visualizes the affinity matrix across iterations for a subset of the YaleB dataset consisting of 10 groups. We observe that after a few hundred iterations, the clusters start forming clearly. This is due to our approach for computing and updating the matrix $P \in \mathbb{R}^{n \times m}$.

\begin{figure*}
  \begin{center}
    \begin{subfigure}[b]{0.3\textwidth}
    \includegraphics[width=0.75\linewidth]{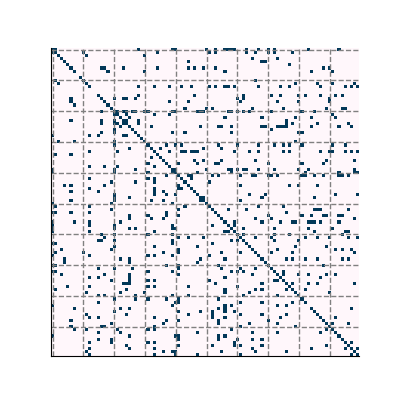}
    \caption{Epoch 0.}
    
  \end{subfigure}
  \begin{subfigure}[b]{0.3\textwidth}
     \includegraphics[width=0.75\linewidth]{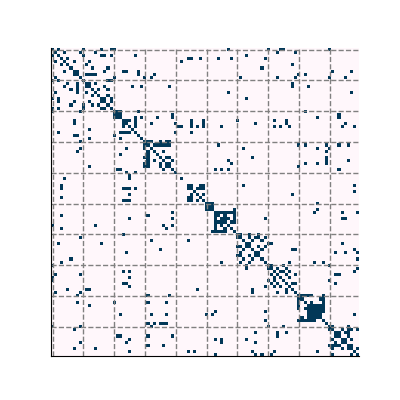}
     \caption{Epoch 100.}
  
  \end{subfigure}
 \begin{subfigure}[b]{0.3\textwidth}
     \includegraphics[width=0.75\linewidth]{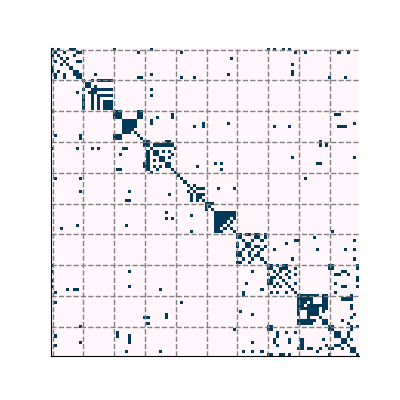}
     \caption{Epoch 500.}
   
  \end{subfigure}
  \caption{Visualizing the graph formed by the affinity matrix on ORL with 10 classes.}
  \label{graph ORL}
  \end{center}
  \vspace{-4mm}
\end{figure*}

Figure \ref{graph ORL} shows the affinity matrix across iterations for a subset of the ORL dataset consisting of 10 groups. This dataset consists of 400 face images, with 40 subjects each having 10 images. Clustering is more challenging due to the increased non-linearity in face subspaces and the small dataset size. We observe that after a few hundred iterations, the clusters begin to form clearly.

\begin{figure*}
  \begin{center}
    \begin{subfigure}[b]{0.3\textwidth}
    \includegraphics[width=0.75\linewidth]{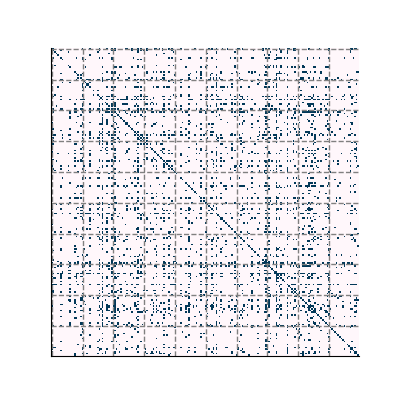}
    \caption{Epoch 0.}
    
  \end{subfigure}
  \begin{subfigure}[b]{0.3\textwidth}
     \includegraphics[width=0.75\linewidth]{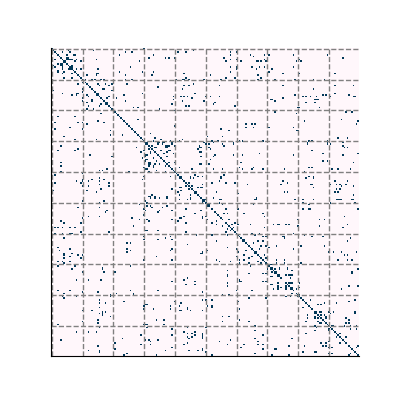}
     \caption{Epoch 100.}
  
  \end{subfigure}
 \begin{subfigure}[b]{0.3\textwidth}
     \includegraphics[width=0.75\linewidth]{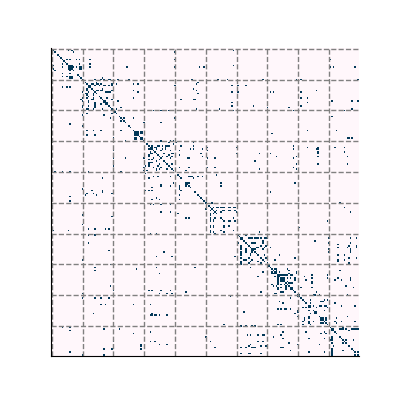}
     \caption{Epoch 500.}
   
  \end{subfigure}
  \caption{Visualizing the graph formed by the affinity matrix on the synthetic dataset with 10 classes.}
  \label{graph Synt10class}
  \end{center}
  \vspace{-4mm}
\end{figure*}

Figure \ref{graph Synt10class} visualizes the evolution of the affinity matrix across training epochs on the synthetic dataset with 10 classes. We observe a progressive emergence of cluster structure similar to that seen in the real datasets (Figures~\ref{graph YaleB} and~\ref{graph ORL}), further validating the convergence behavior of our model.

\begin{figure*}
  \begin{center}
    \begin{subfigure}[b]{0.3\textwidth}
    \includegraphics[width=0.75\linewidth]{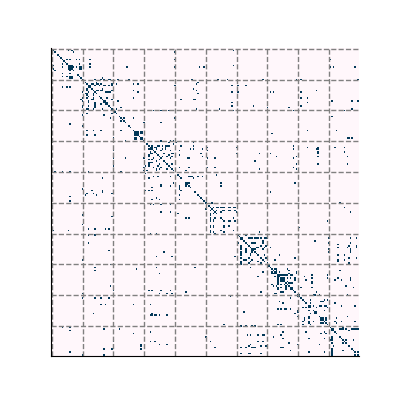}
    \caption{200 points.}
    
    \end{subfigure}
        \begin{subfigure}[b]{0.3\textwidth}
     \includegraphics[width=0.75\linewidth]{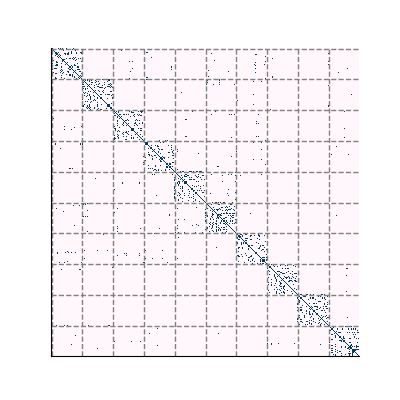}
     \caption{300 points.}
     
    \end{subfigure}
     \begin{subfigure}[b]{0.3\textwidth}
     \includegraphics[width=0.75\linewidth]{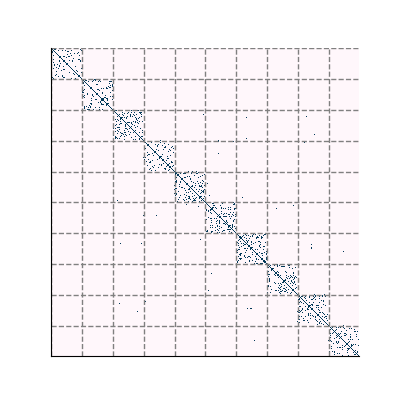}
     \caption{500 points.}
     
          \end{subfigure}
  
  \caption{Visualizing the graph formed by the affinity matrix on different sizes of the synthetic dataset for 500 epochs.}
  \label{graph Synt}
  \end{center}
  \vspace{-4mm}
\end{figure*}

Figure \ref{graph Synt} presents the affinity matrix for different dataset sizes consisting of 10 groups, from the synthetic dataset, for a training run of 500 iterations. We observe that as the dataset size increases, the clusters become more distinct. Experiments on synthetic data confirm the ability of our approach to effectively identify subspaces.

\section{Conclusion}

In this paper, we presented SDSNet, a scalable deep subspace clustering framework designed to address the computational challenges of traditional subspace clustering methods. By integrating deep learning with a novel, scalable formulation, SDSNet reduces the complexity of affinity matrix construction and spectral clustering from \(O(n^3)\) to \(O(n)\). The proposed method leverages landmark points to approximate the affinity matrix, enabling efficient clustering without sacrificing accuracy. Additionally, the use of a deep auto-encoder allows SDSNet to handle complex, nonlinear subspaces, further enhancing its clustering performance. Across five datasets, SDSNet yields strong clustering results. It attains the top ACC and NMI on YaleB and UMIST, remains strong on ORL (best ACC), and is close to the best elsewhere. Furthermore, SDSNet achieves competitive results compared to non-scalable deep subspace clustering methods, highlighting its ability to balance computational efficiency and clustering performance. The scalability analysis confirms that SDSNet maintains linear complexity with respect to dataset size, making it a practical solution for scalable subspace clustering.


\balance


\end{document}